          [2001/04/25 1.1 (PWD)]
\documentclass[a4paper,twocolumn]{esapub2005} 
\usepackage{times}
\usepackage{url}
\usepackage{graphicx}
\usepackage{multirow}
\usepackage[dvipsnames]{xcolor}
\usepackage{siunitx}
\usepackage{gensymb}
\usepackage[font={small, it}]{caption} 
\usepackage{titlesec}
    \titlespacing*{\section}{0pt}{0.5ex}{0.3ex} 
    \titlespacing*{\subsection}{0pt}{0.3ex}{0.2ex} 

\title{Fast Vision in the Dark: A Case for Single-Photon Imaging in Planetary Navigation}
\author{David Rodríguez-Martínez$^{*}$}
\author{C.J. Pérez del Pulgar}
\affil{Space Robotics Lab, Dept. of Systems Engineering and Automation, University of Malaga, \\ $^{*}$corresponding author: david.rm@uma.es}

\begin{document}


\maketitle

\begin{abstract}
Improving robotic navigation is critical for extending exploration range and enhancing operational efficiency. Vision-based navigation relying on traditional CCD or CMOS cameras faces major challenges when complex illumination conditions are paired with motion, limiting the range and accessibility of mobile planetary robots. In this study, we propose a novel approach to planetary navigation that leverages the unique imaging capabilities of Single-Photon Avalanche Diode (SPAD) cameras. We present the first comprehensive evaluation of single-photon imaging as an alternative passive sensing technology for robotic exploration missions targeting perceptually challenging locations, with a special emphasis on high-latitude lunar regions. We detail the operating principles and performance characteristics of SPAD cameras, assess their advantages and limitations in addressing key perception challenges of upcoming exploration missions to the Moon, and benchmark their performance under representative illumination conditions. 
\end{abstract}

\section{Introduction}

\begin{figure*}[ht]
    \centering
    \includegraphics[width=0.9\textwidth]{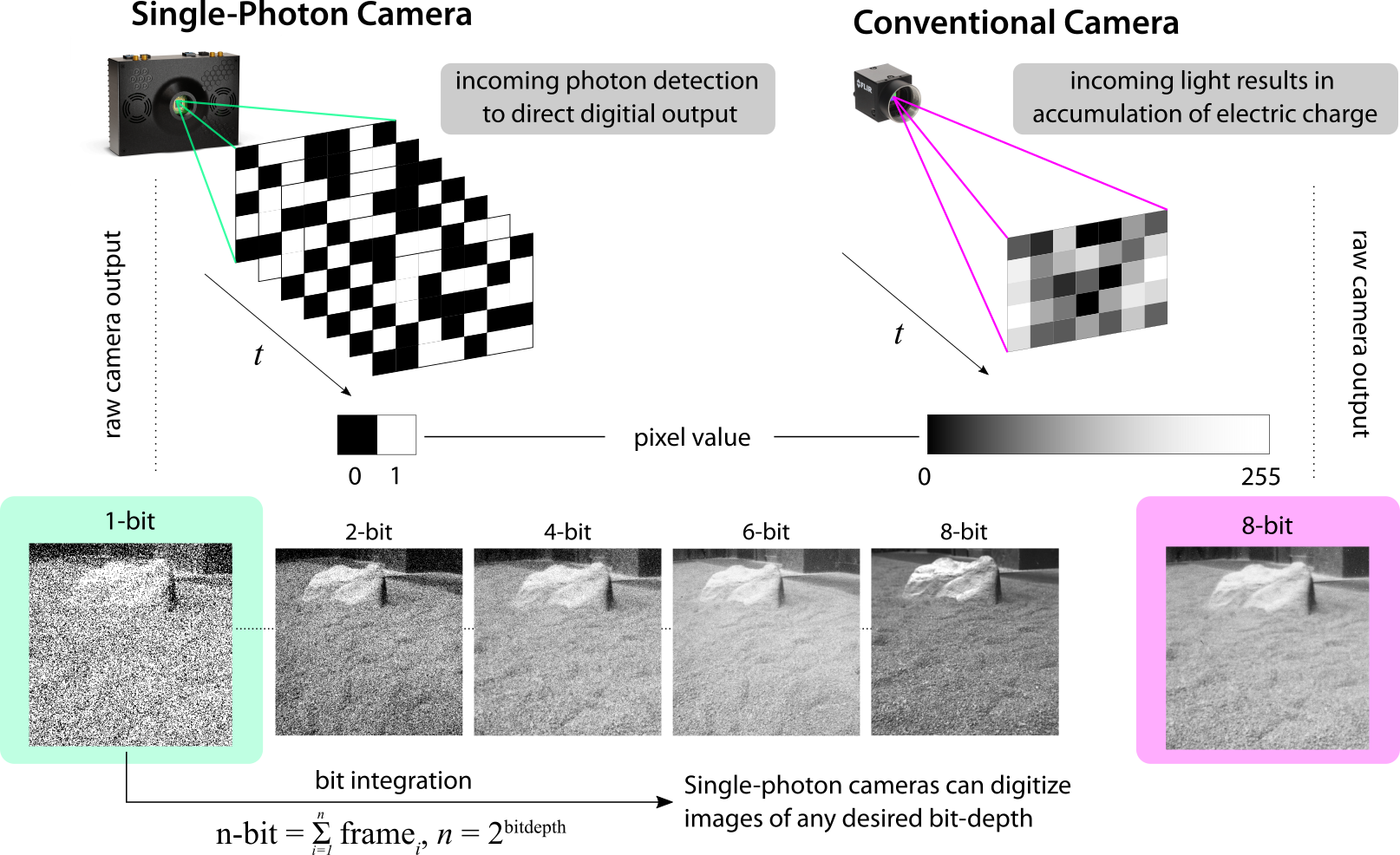}
    \caption{Unlike conventional cameras, which accumulate electric charge in each pixel over a given amount of time and subsequently convert it to a digital value (often of 8 bits per pixel), single-photon cameras provide a direct and synchronized digital output in the form of binary frames (1 bit per pixel). Each pixel value (0 or 1) indicates the absence or presence of an incident photon. This photon-level sensitivity enables the generation of images with arbitrary bit-depth by integrating 2$^n$ number of binary frames, where $n$ is the desired bit-depth of the final image. SPADs can also capture these binary frames at very high speeds, up to 100\,kfps (10\,$\mu$  per frame), or even combined binary frames taken at different exposures to create high dynamic range images (\textgreater 100\,dB).}
    \label{fig:spad_intro}
\end{figure*}

Enhanced sensing and imaging systems are essential to the success of a number of high-priority planetary exploration missions, particularly those targeting environments where visual perception is severely impaired. Examples include prospecting at high-latitude lunar locations, characterized by low solar elevation angles and shorter diurnal cycles, or navigating the pitch-black interiors of Martian lava tubes. These environments present extreme illumination conditions---ranging from partial to complete darkness---paired with uneven, hazard-strewn landscapes. Navigating autonomously in such conditions demands high visual acuity, while also conserving limited onboard power and computational resources. 

Vision-based navigation relying on Charge-Coupled Device (CCD) and Complementary Metal-Oxide-Semiconductor (CMOS) cameras faces significant challenges under complex and dynamic lighting conditions, which limit the operational range and accessibility of planetary robots. For example, martian rovers can typically drive for only about 3 hours per sol when illumination conditions are optimal \cite{verma2023}. Similarly, NASA's \textit{Ingenuity} helicopter was unable to fly when atmospheric conditions on Mars were best suited for it (dawn and dusk), since the insufficient illumination prevented its visual odometry from reliably detecting and matching ground features \cite{bayard2019}. 

To address the lack of illumination, inconsistent brightness, and faster-than-usual traveling speeds \cite{rodriguez2019high, trimble2023viper}, high-latitude lunar exploration and prospecting missions must rely on either more robust computational image-processing algorithms (e.g., \cite{soussa2025luvo}), expand exteroceptive sensor suites to include a wider range of imaging modalities \cite{vidal2018}, employ improved and more reliable sensing technologies \cite{mahlknecht2022exploring, polizzi2022}, or, most likely, a combination thereof. 

Among alternative sensing technologies, neuromorphic (event-based) cameras \cite{mahlknecht2022exploring}, thermal-infrared sensors \cite{castilla2023thermal}, and active scanning Light Detection and Ranging (LiDAR) sensors \cite{christian2013survey} are presented as promising alternatives for enabling robust navigation in perceptually challenging environments, outperforming conventional optics in dimly lit, high dynamic range (HDR), and fast-moving scenarios. LiDAR, in particular, has already presented promising results for precise mapping of celestial body surfaces \cite{smith2001mars, hussmann2018bepicolombo}, spacecraft navigation and landing \cite{leonard2022cross}, and autonomous in-orbit rendezvous and docking maneuvers \cite{pyrak2022performance}. But their application and proven performance in planetary exploration remains limited, largely due to the current, though rapidly maturing, technology readiness of solid-state, flash, and global shutter LiDARs \cite{christian2013survey}. Additional constraints include high power demands, mass and size constraints, limited range and restricted fields of view, and substantial data throughput requirements, which together demand careful engineering trade-offs before such systems can be reliably integrated into planetary missions. Notably, what characterizes all of the aforementioned alternative technologies is their distinct way of parsing reality, making it often necessary to pair them with complementary optical sensors to provide the complete contextual information most missions demand. 

In this work, we introduce a novel imaging technology with a unique potential to enhance vision-based planetary robotic systems due to its light sensitivity, speed, adaptability, and, despite its distinctive way of capturing visual information, its ability to operate and record visual data without the need for complementary sensors: \textbf{single-photon imaging}. Our main contributions include:

\begin{enumerate}
    \item \textbf{Introduction of single-photon imaging for planetary navigation}. We focus on Single-Photon Avalanche Diode (SPAD) cameras, summarizing their operating principles, modeling methods, key characteristics, and assessing their advantages and limitations in addressing critical perception challenges of upcoming exploration missions. 
    \item \textbf{First evaluation of SPAD cameras as a passive imaging alternative for planetary navigation}. We compare their imaging and task-specific performance with a standard monochrome camera under lunar-analog conditions replicating the visual complexity and variable illumination of high-latitude regions.
    \item \textbf{Identification of key challenges and future research directions}. We outline key obstacles to increasing SPAD technology readiness and enabling its integration into future missions, as well as potential benefits and unexplored research avenues.
\end{enumerate}

\section{Single-photon imaging}

\subsection{Operating Principle}

Single-Photon Avalanche Diode (SPAD) cameras are a specific type of avalanche photo-diodes (APDs) capable of detecting the arrival of individual photons. They do so by operating in the so-called Geiger mode; i.e., at a high reverse bias voltage (above breakdown) where the strike of a single photon triggers an avalanche multiplication of charge carriers in the semiconductor, producing a distinct electrical pulse as an output \cite{swissSPAD}. This pulse is amplified directly within the SPAD pixel, providing a ready-to-use, inherently digital signal in the form of a synchronous binary stream where 0 means no photon detected and 1 means a photon was detected during that time window (see Figure~\ref{fig:spad_intro}). This capability, together with active quenching (rapid lowering of the bias voltage below breakdown to stop the otherwise self-sustaining avalanche process), enables SPADs to detect photon strikes with timing resolutions in the sub-10 picosecond range \cite{gramuglia2022a} and exceptionally low noise levels \cite{gramuglia2022b}. Note that, unlike other forms of photon-detection such as Quanta Image Sensors (QIS) based on extremely small specialized pixels called jots \cite{fossum2016quanta}, SPADs are capable of detecting single-photon events but cannot resolve the exact number of photons arriving simultaneously within each detection time window \cite{ma2022ultra}. 

The implementation of SPADs in CMOS technology \cite{charbon2014} has facilitated their miniaturization and integration into large pixel arrays (\textgreater 1Mpx, e.g., \cite{morimoto2020, morimoto20213}). This advancement has enabled their use as extremely fast (up to 100\;kpfs) and light-sensitive intensity imaging sensors, making SPADs suitable for a variety of applications, including robotics. Unlike conventional cameras, SPADs produce a noiseless, constant stream of binary frames. These frames can be accumulated and integrated to produce images of any desired bit depth (Figure~\ref{fig:spad_intro}). Even binary frames captured at different exposure times can be combined to produce high-speed HDR images. 

These capabilities enable resolving visual features without the need for artificial lighting or auxiliary sensors, improving adaptability and robustness under extreme conditions (e.g., mlx illuminance) while optimizing system design trade-offs. SPADs are already being used in a wide variety of applications, from 2D picosecond-resolution biomedical imaging \cite{zickus2020} to active 3D time-of-flight (ToF) sensing and LiDAR \cite{tsai2018spad}. Their rapid development has also drawn attention in the computer vision community (\cite{liu2022single, ma2023burst, koerner2024photon}, to name a few). But their application and potential added-value have been seldom explored in the context of robotics and space \cite{wu2022radiation, guerrieri2024dynamically}. 

\subsection{Imaging Model}

The arrival of photons at a single SPAD pixel can be modeled as a random process. Assuming constant illuminance (and photon independence), the probability of a number of photons, $k$, striking a pixel within an exposure window follows a Poisson distribution \cite{yang2011bits}: 

\vspace{-1em}
\begin{equation}
    P(x=k) = \frac{\lambda^k e^{-\lambda}}{k!},
\end{equation}

where $\lambda$ is the expected number of photons, determined by the photon flux, $\phi$ (photons/second), the sensor quantum efficiency, $\eta$, and the exposure time, as in $\tau$: $\lambda = E(x) = \phi \eta \tau$. If additional effects, such as dark counts (false detections in SPADs due to thermal noise triggering avalanches without the presence of incident photons), $\lambda$ can also be defined as $\lambda = (\phi \eta + r_d)\tau$, where $r_d$ indicates the dark count rate (DCR) in counts per second per pixel. In conventional sensors, this Poisson-distributed count directly determines the pixel intensity. For SPADs, however, the binary detection nature of every exposure means each pixel cannot count beyond 1 in a single frame, returning as a result a value of 1 if $k\geq1$ or a value of 0 if $k=0$. This detection is, therefore, governed by discrete Bernoulli statistics, where:

\vspace{-1em}
\begin{eqnarray}
    P(x=0) =  e^{-\lambda} = e^{-(\phi \eta \tau)}\\
    P(x=1) = 1 - P(x=0) = 1 - e^{-(\phi \eta \tau)}
\end{eqnarray}

Over $N$ binary frames (assuming no motion and perfect alignment), the maximum likelihood estimate (MLE) of $P(x=1)$ per frame is simply defined by $\hat{p}=n/N$, where $n$ is the number of frames with $k\geq1$. One can, therefore, estimate the MLE of the photon arrival rate per pixel, $\hat{\phi}$, as:

\vspace{-1em}
\begin{equation}
    \hat{\phi} = \frac{1}{\eta \tau}(-ln(1 - \hat{p})) = \frac{1}{\eta \tau}(-ln(1 - \frac{n}{N})). 
\end{equation}

This statistical framework enables intensity reconstruction from sequences of binary frames.

\section{Methodology}

We grouped our evaluation of the performance of a single-photon camera against that of a standard monochrome camera into two categories. First, we assessed the \textbf{imaging performance} of both cameras under a range of illumination conditions---dawn/dusk, noon, and night---corresponding to illuminance levels from 100\,lx down to 0.1\,lx, at equivalent camera exposure settings. These conditions were specifically designed to simulate lighting conditions at the lunar poles, where Sun elevation angles range between 4\degree and 8\degree during daylight. Second, we assessed the \textbf{task-specific performance} of both cameras on mission-relevant tasks, including (A) scene segmentation, (B) lander detection, and (C) rover localization. For the first two, performance is evaluated over the results of the Segment Anything Model (SAM) \cite{kirillov2023segment}, a transformer-based segmentation framework capable of performing zero-shot object and region segmentation across diverse image domains without task-specific training. We specifically implemented the pre-trained ViT-H model. For localization, we run our data on the well-known ORB-SLAM3 \cite{campos2021orb} pipeline, a feature-based, real-time visual-inertial SLAM system that supports monocular, stereo, and inertial input. These evaluations allowed us to compare not only raw imaging quality but also the practical impact of each camera on key operational tasks.

\begin{figure}[htb]
    \centering
    \includegraphics[width=\linewidth]{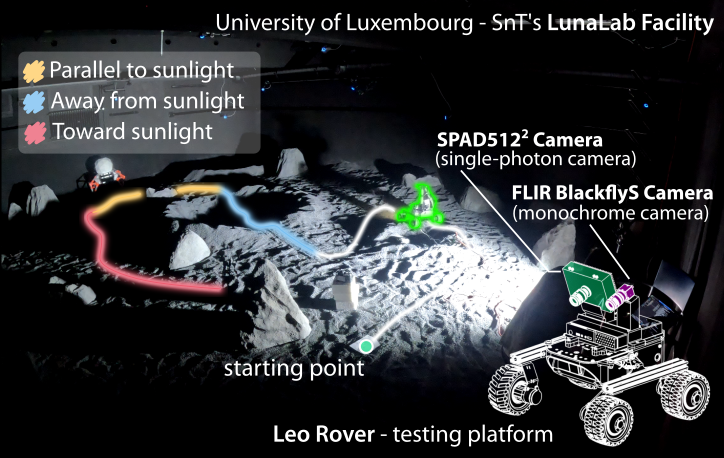}
    \caption{Experimental setup for our tests at the LunaLab.}
    \label{fig:methodology}
\end{figure}

\subsection{Sensors}

We used a SPAD512 single-photon camera from Pi Imaging Technology and a FLIR Blackfly-S from Teledyne Vision Solutions as our standard monochrome camera. Although the Blackfly-S is not a space-qualified camera, it is widely used in robotic vision applications, making it a relevant baseline for comparison. Optics for both cameras were chosen to account for their respective sensor sizes, ensuring similar equivalent focal lengths. The key specifications of each sensor are summarized in Table~\ref{tab:cameras}. Both cameras were mounted on a Leo Rover testing platform as shown in Figure~\ref{fig:methodology}. The rover also incorporated headlights to artificially illuminate the scene. 

\begin{table}
  \begin{center}
    \caption{Conventional vs single-photon camera specifications.}\vspace{-0.5em}
    \begin{tabular}[h]{lcc}
      \hline
       & FLIR Blackfly-S   &  SPAD 512 \\
      \hline
      Sensor size & 6.23$\times$4.98\,mm & 9.5$\times$9.5\,mm \\
      Resolution & 720$\times$540 & 512$\times$512\\
      Pixel Pitch & 6.9\,\SI{}{\micro\metre} & 16.38\,\SI{}{\micro\metre}\\
      Bit Depth (raw) & 8-bit & 1-bit\\
      EFL & 14.24\,mm f1.4 & 14.88\,mm f1.4\\
      FoV & 21.6\degree$\times$12.3\degree & 35.5\degree$\times$35.5\degree\\
      \hline 
      \multicolumn{3}{l}{\footnotesize{EFL: Effective Focal Length; FoV: Field of View}} \\
      \end{tabular}
    \label{tab:cameras}
  \end{center}
\end{table}

\subsection{Data}

The images used in our evaluation can all be found in the SPICE-HL3 dataset \cite{rodriguez2025spice}, recorded at the LunaLab of the University of Luxembourg-SnT. This facility is designed to replicate, with high photometric fidelity, lunar lighting conditions at various latitudes (Figure~\ref{fig:methodology}). The facility uses an Aputure LightStorm 600c Pro 600W spotlight, whose vertical and horizontal position, color temperature (2,300–10,000 K), and illuminance (69–51,100 lx) can be precisely adjusted to simulate different solar elevation angles and daylight conditions.

For the imaging performance evaluation, we primarily use images from the SPICE-HL3 static trajectory A. Data are divided into three main sections based on the heading angle with respect to the sun-simulator spotlight. These are referred to as \textit{parallel to sunlight},\textit{ away from sunlight}, and \textit{toward sunlight} throughout the text. Visual odometry performance is evaluated on trajectories B, C, and D, each of which also corresponds to each of the previous phase angles, with the rover moving at 0.5 m/s. Because the native image resolutions of the SPAD and conventional camera differ, images are cropped to match FLIR's aspect ratio of 4:3 for imaging performance, scene segmentation, and lander detection, and the SPAD's 1:1 aspect ratio for visual odometry. To ensure comparability and compatibility with the implemented algorithms, most evaluations use SPAD images digitized at 8-bit. Although for the scene segmentation, we additionally look into the performance of the SPAD at a lower bit depth of 4 bits per pixel. No image enhancement or additional preprocessing is applied to either the SPAD or the FLIR data. 

\subsection{Metrics}

Imaging performance was assessed using both qualitative and quantitative measures. Quantitative evaluation included contrast (RMS of pixel intensities), entropy (randomness of the intensity distribution, with higher values indicating greater detail), and sharpness (based on the variance of the Laplacian response). We also computed BRISQUE and MS-SSIM scores. BRISQUE, or Blind/Referenceless Image Spatial Quality Evaluator, is a no-reference metric that models natural scene statistics to estimate perceived distortion (lower scores indicate better quality). MS-SSIM, or Multi-Scale Structural Similarity Index, is a full-reference structural similarity measure that operates across multiple resolutions (score range 0–1, with 1 indicating perfect agreement with a ground truth image). For MS-SSIM, images captured under regular laboratory illumination served as the reference.

For scene segmentation and lander identification, we evaluated the mean Intersection over Union (mIoU), inference time, and number of masks generated, with the latter two providing preliminary insights into computational demand variations across input images. In the visual odometry experiments, we measured the RMSE of the Absolute Trajectory Error (ATE) and the maximum drift (maximum ATE) for each trajectory.

\section{Results and discussion}

\subsection{Imaging performance}

\begin{figure*}[htb]
    \centering
    \includegraphics[width=\textwidth]{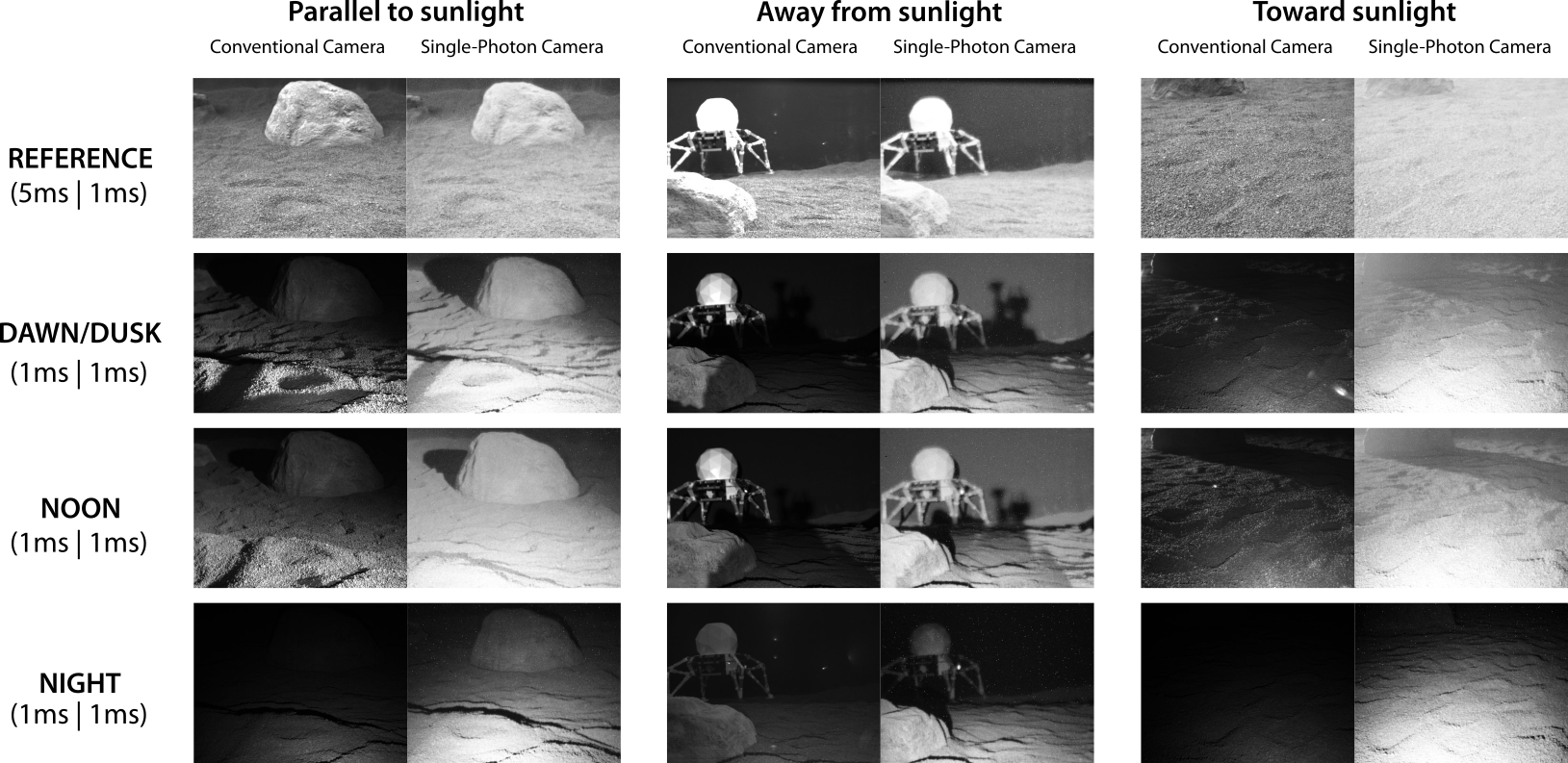}
    \caption{Selected frames taken by a conventional camera and a single-photon camera under different illumination conditions and at 8-bit equivalent exposure times. These frames are extracted from different sections of the SPICE-HL3 dataset trajectory A \cite{rodriguez2025spice}. All images were captured with the rover headlights switched on.}
    \label{fig:quality}
\end{figure*}

The first step of our evaluation involved a direct visual assessment of the captured images to identify qualitative differences between the two camera types. Figure~\ref{fig:quality} presents example frames from each of the three distinct sections along trajectory A at different illumination conditions. All images evaluated were acquired with the rover headlights switched on. For the reference images, exposure times were chosen to yield well-balanced histogram distributions (i.e., mean intensity values near the midpoint of the dynamic range without clipping pixels on either edge). For the remaining images, exposure times were selected from those recorded under identical settings for both cameras, ensuring a fair basis for comparison.

The single-photon camera appears to consistently capture more detail of the scene under low-light conditions compared to the conventional camera. Under reference illumination, both camera types produce comparable images, with the conventional camera presenting a higher contrast and sharpness, likely due to its slightly higher resolution and efficiency at large photon counts. But as illumination decreases, the conventional camera shows a significant loss of detail, particularly at night, requiring a much longer exposure to resolve equal levels of detail in the image. In contrast, the single-photon camera maintains visibility of ground textures and obstacle features across all lighting conditions, capturing a better balance between lights and shadows, despite a clear lack of contrast. 

Quantitative evaluation of image quality metrics across illumination conditions and exposure times highlights clear performance differences between the two camera types. The single-photon camera consistently outperforms the conventional camera in low-light scenarios (dawn/dusk and night) across most metrics, often achieving superior results at equal or substantially shorter exposure times. This advantage persists regardless of artificial illumination, underscoring the SPAD sensor’s reduced reliance on external light sources. Nighttime results amplify this trend, where SPAD images exhibit higher contrast, entropy, and sharpness. In brighter conditions (noon) and at longer exposures, however, the performance gap narrows, with the conventional camera matching or surpassing SPAD performance in some metrics. At these longer exposures, the conventional camera often delivers the highest metric values across the full range of illumination conditions and exposure settings. This advantage likely arises from a combination of factors. The FLIR’s higher pixel count enables it to capture finer spatial detail when sufficient photons are available. In bright, static scenes, longer exposure times allow conventional cameras (often with higher quantum efficiency and better matching sensor-optics integration) to collect significantly more photons per pixel, increasing the signal-to-noise ratio (SNR) and reducing shot noise. In contrast, the SPAD camera may reach an SNR plateau earlier due to saturation of its pixels. Note the SPAD512 model used in this evaluation did not incorporate microlenses (limiting the fill factor to $\approx$ 10\%), resulting in limited photon detection efficiencies (PDEs), often below 50\%. This highlights the disparities in the degree of maturity of both technologies, with conventional camera sensors benefiting from a level of technology development that single-photon sensors have yet to achieve. SPAD technology, by design, is optimized for low-light, high-speed acquisition rather than maximizing image quality in static, high photon flux conditions. Overall, these results affirm SPAD's efficacy for low-latency, high-fidelity capture in challenging illuminations, while also highlighting context-dependent scenarios in which conventional imaging technology yet retains an advantage.

\vspace{-0.5em}
\begin{table*}[t]
\small  
\centering
\caption{Comparison between a conventional camera (FLIR Blackfly-S) and a single-photon camera (SPAD512) across multiple metrics. Values are reported in the order FLIR $\mid$ SPAD and represent the average across all frames in each sequence. SPAD exposure times are given as 8-bit equivalents. Contrast and sharpness values are normalized. Highlighted entries indicate the best value for each combination of illumination condition and exposure time. The overall best values across the exposure range are highlighted in green for SPAD and in pink for FLIR. An $\times$ denotes failed image acquisitions due to either insufficient light or excessive saturation.}\vspace{-0.5em}
\begin{tabular}{cccccccc}
\hline
Illumination      & HL & Exposure (ms)  & Contrast     & Entropy       & Sharpness    & BRISQUE & MS-SSIM \\ \hline
\multirow{13}{*}{Dawn/Dusk}  & \multirow{6}{*}{Off}        & $\times$ $\mid$ 0.128       & $\times$ $\mid$ \textbf{0.0833}   & $\times$ $\mid$ \textcolor{Green}{\textbf{0.417}}    & $\times$ $\mid$ \textbf{0.006}    & $\times$ $\mid$ \textbf{41.06} & $\times$ $\mid$ \textbf{0.047} \\
                             &         & $\times$  $\mid$ 0.256  & $\times$  $\mid$ \textbf{0.101 }& $\times$  $\mid$ \textbf{0.344 } & $\times$  $\mid$ \textbf{0.009}  & $\times$  $\mid$ \textbf{40.62} & $\times$ $\mid$ \textbf{0.059} \\
                             &         & 1.00 $\mid$ 1.28     & 0.085 $\mid$ \textbf{0.153} & 0.044 $\mid$ \textbf{0.055}  & 0.014 $\mid$ \textbf{0.018}   & 51.11 $\mid$ \textbf{39.70} & 0.026 $\mid$ \textbf{0.083} \\
                             &         & 2.50 $\mid$ 2.56     & 0.157 $\mid$ \textbf{0.181} & \textbf{0.017} $\mid$ 0.016  & \textbf{0.030} $\mid$ 0.025  & 46.46 $\mid$ \textbf{38.74} & 0.052 $\mid$ \textbf{0.094}\\
                             &         & 10.00 $\mid$ 12.8     & 0.215 $\mid$ \textcolor{Green}{\textbf{0.229}} & 0.00 $\mid$ \textbf{0.063}  & 0.029 $\mid$ \textcolor{Green}{\textbf{0.073}}  & \textcolor{VioletRed}{\textbf{38.08}} $\mid$ 42.59 & \textcolor{VioletRed}{\textbf{0.153}} $\mid$ 0.033 \\
                             &         & 50.00 $\mid$ $\times$ & \textbf{0.182} $\mid$ $\times$ & 0.00 $\mid$ $\times$ & \textbf{0.0113} $\mid$ $\times$  & $\times$ $\mid$ $\times$ & \textbf{0.3018} $\mid$ $\times$ \\ 
                             & \multirow{6}{*}{On}          & $\times$ $\mid$ 0.128    & $\times$ $\mid$ \textbf{0.177}   & $\times$ $\mid$ \textcolor{Green}{\textbf{0.090}}    & $\times$ $\mid$ \textbf{0.017}    & $\times$ $\mid$ \textbf{39.23} & $\times$ $\mid$ \textbf{0.029}    \\
                             &          & $\times$ $\mid$ 0.256     & $\times$ $\mid$ \textbf{0.218} & $\times$ $\mid$ \textbf{0.050}  & $\times$ $\mid$ \textbf{0.020}  & $\times$ $\mid$ \textbf{38.02} & $\times$ $\mid$ \textbf{0.033}  \\
                             &          & 1.00 $\mid$ 1.28     & 0.137 $\mid$ \textbf{0.289} & \textbf{0.009 } $\mid$  0.00 & 0.016 $\mid$ \textbf{0.02}  & \textbf{40.03} $\mid$ 41.23 & 0.029 $\mid$ \textbf{0.058} \\
                             &          & 2.50 $\mid$ 2.56     & 0.244 $\mid$ \textbf{0.273} & \textbf{0.002} $\mid$ 0.00  & \textcolor{VioletRed}{\textbf{0.031}} $\mid$ 0.016  & \textbf{35.08} $\mid$ 37.56 & 0.054 $\mid$ \textbf{0.056} \\
                             &          & 10.00 $\mid$ 12.80     & \textcolor{VioletRed}{\textbf{0.296}} $\mid$ 0.146 & 0.00 $\mid$ \textbf{0.022}  & \textbf{0.027} $\mid$ 0.024  & \textcolor{VioletRed}{\textbf{33.19}} $\mid$ 48.16 & \textcolor{VioletRed}{\textbf{0.066}} $\mid$ 0.022 \\
                             &         & 50.00 $\mid$ $\times$ & \textbf{0.181} $\mid$ $\times$ & 0.00 $\mid$ $\times$  & \textbf{0.007} $\mid$ $\times$ & \textbf{44.71} $\mid$ $\times$ & \textbf{0.055} $\mid$ $\times$ \\\hline
\multirow{13}{*}{Noon}       & \multirow{6}{*}{Off}         & $\times$  $\mid$ 0.128    & $\times$  $\mid$ \textbf{0.109}   & $\times$  $\mid$ \textcolor{Green}{\textbf{0.241}}   & $\times$ $\mid$ \textbf{0.009}    & $\times$ $\mid$ \textbf{40.57} & $\times$  $\mid$ \textbf{0.02} \\
                             &         & $\times$  $\mid$ 0.256     & $\times$  $\mid$ \textbf{0.130} & $\times$  $\mid$ \textbf{0.163}  & $\times$  $\mid$ \textbf{0.012}  & $\times$  $\mid$ \textbf{39.64}  & $\times$  $\mid$ \textbf{0.037} \\
                             &         & 1.00 $\mid$ 1.28     & 0.122 $\mid$ \textbf{0.182} & \textbf{0.036} $\mid$ 0.0154  & \textbf{0.021} $\mid$ 0.017  & 45.26 $\mid$ \textbf{38.50} & 0.093 $\mid$ \textbf{0.148} \\
                             &         & 2.50 $\mid$ 2.56     & \textbf{0.220} $\mid$ 0.206 & \textbf{0.011} $\mid$ 0.0017  & \textbf{0.047} $\mid$ 0.023  & 41.14 $\mid$ \textbf{37.91} & 0.117 $\mid$ \textbf{0.148} \\
                             &         & 10.00 $\mid$ 12.80     & \textcolor{VioletRed}{\textbf{0.279}} $\mid$ 0.209 & 0.00 $\mid$ \textbf{0.059}  & \textcolor{VioletRed}{\textbf{0.043}} $\mid$ 0.041  & \textcolor{VioletRed}{\textbf{33.63}} $\mid$ 43.37 & \textbf{0.146} $\mid$ 0.010 \\
                             &         & 50.00 $\mid$ $\times$ & \textbf{0.188} $\mid$ $\times$ & 0.00 $\mid$ $\times$  & \textbf{0.010} $\mid$ $\times$  & \textbf{39.91} $\mid$ $\times$ & \textcolor{VioletRed}{\textbf{0.192}} $\mid$ $\times$ \\
                             & \multirow{6}{*}{On}          & $\times$ $\mid$ 0.128       & $\times$ $\mid$ \textbf{0.173}   & $\times$ $\mid$ \textcolor{Green}{\textbf{0.042}}    & $\times$ $\mid$ \textbf{0.018}    & $\times$ $\mid$ \textbf{40.61}  & $\times$ $\mid$ \textbf{0.015}  \\         
                             &          & $\times$ $\mid$ 0.256    & $\times$ $\mid$ \textbf{0.207} & $\times$ $\mid$ \textbf{0.014}  & $\times$ $\mid$ \textbf{0.020}  & $\times$ $\mid$ \textbf{38.86} & $\times$ $\mid$ \textbf{0.100} \\
                             &          & 1.00 $\mid$ 1.28     & 0.1493 $\mid$ \textbf{0.249} & \textbf{0.005} $\mid$ 0.00  & \textbf{0.022} $\mid$ 0.017  & \textbf{38.82} $\mid$ 41.08 & 0.087 $\mid$ \textbf{0.096} \\
                             &          & 2.50 $\mid$ 2.56     & \textbf{0.255} $\mid$ 0.226 & \textbf{0.0009} $\mid$ 0.00  & \textcolor{VioletRed}{\textbf{0.041}} $\mid$ 0.013 & \textbf{35.85} $\mid$ 38.51 & \textcolor{VioletRed}{\textbf{0.114}} $\mid$ 0.084 \\
                             &          & 10.00 $\mid$ 12.80     & \textcolor{VioletRed}{\textbf{0.268}} $\mid$ 0.093 & 0.00 $\mid$ \textbf{0.026}  & 0.0271 $\mid$ \textbf{0.032}  & \textcolor{VioletRed}{\textbf{34.92}} $\mid$ 47.71 & 0.087 $\mid$ \textbf{0.009}  \\
                             &         & 50.00 $\mid$ $\times$ & \textbf{0.144} $\mid$ $\times$ & 0.00 $\mid$ $\times$  & \textbf{0.005} $\mid$ $\times$  & \textbf{46.67} $\mid$ $\times$ & \textbf{0.038} $\mid$ $\times$ \\\hline
\multirow{13}{*}{Night}      & \multirow{5}{*}{Off}         & 1.28 $\mid$ 1.28       & 0.003 $\mid$ \textbf{0.036}   & 0.00 $\mid$ \textbf{0.171}    & 0.00 $\mid$ \textbf{0.019}    & $\times$ $\mid$ \textbf{45.72}  & 0.020 $\mid$ \textbf{0.049} \\
                             &         & 2.56 $\mid$ 2.56     & 0.0095 $\mid$ \textbf{0.044} & 0.0006 $\mid$ \textcolor{Green}{\textbf{0.261}}  & 0.0001 $\mid$ \textbf{0.029}  & $\times$ $\mid$ \textbf{44.67}  & \textbf{0.070} $\mid$ 0.040 \\
                             &         & 12.80 $\mid$ 12.80     & 0.0361 $\mid$ \textbf{0.074} & 0.00 $\mid$ \textbf{0.010}  & 0.00 $\mid$ \textbf{0.068}  & \textbf{46.12} $\mid$ $\times$  & \textbf{0.269} $\mid$ 0.033 \\
                             &         & 128.00 $\mid$ 128.00     & \textcolor{VioletRed}{\textbf{0.114}} $\mid$ 0.106 & 0.00 $\mid$ \textbf{0.020}  & 0.007 $\mid$ \textcolor{Green}{\textbf{0.142}}  & \textcolor{VioletRed}{\textbf{36.23}} $\mid$ 40.82  &  \textcolor{VioletRed}{\textbf{0.404}} $\mid$ 0.025 \\
                             &         & 307.20 $\mid$ 307.20     & 0.036 $\mid$ \textbf{0.104} & 0.00 $\mid$ \textbf{0.0200}  & 0.00 $\mid$ \textbf{0.135}  & $\times$ $\mid$ \textbf{42.97}  & 0.044 $\mid$ \textbf{0.113} \\
                             & \multirow{6}{*}{On}          & $\times$ $\mid$ 0.128       & $\times$ $\mid$ \textbf{0.170}  & $\times$ $\mid$ \textcolor{Green}{\textbf{0.387}}    & 5$\times$ $\mid$ \textbf{0.014}    & $\times$ $\mid$ \textbf{43.51}   & $\times$ $\mid$ \textbf{0.005}  \\
                             
                             &         & $\times$ $\mid$ 0.256     & $\times$ $\mid$ \textbf{0.220} & $\times$ $\mid$ \textbf{0.247}  & $\times$ $\mid$ \textbf{0.018}  & $\times$ $\mid$ \textbf{43.09}  & $\times$ $\mid$ \textbf{0.007} \\
                             &          & 1.00 $\mid$ 1.28     & 0.098 $\mid$ \textbf{0.339} & 0.016 $\mid$ \textbf{0.030}  & 0.002 $\mid$ \textbf{0.023}  & 43.99 $\mid$ \textbf{42.65}  & 0.021 $\mid$ \textbf{0.030} \\
                             &          & 2.50 $\mid$ 2.56     & 0.207 $\mid$ \textcolor{Green}{\textbf{0.353}} & \textbf{0.006} $\mid$ 0.0017  & 0.010 $\mid$ \textbf{0.021}  & 40.67 $\mid$ \textbf{39.66}  & \textbf{0.058} $\mid$ 0.037 \\
                             &          & 10.00 $\mid$ 12.80     & \textbf{0.319} $\mid$ 0.291 & 0.00 $\mid$ \textbf{0.058}  & 0.021 $\mid$ \textcolor{Green}{\textbf{0.070}}  & \textcolor{VioletRed}{\textbf{36.35}} $\mid$ 47.06  & \textbf{0.079} $\mid$ 0.045 \\
                             &         & 50.00 $\mid$ $\times$ & \textbf{0.231} $\mid$ $\times$ & 0.00 $\mid$ $\times$  & \textbf{0.007} $\mid$ $\times$  & \textbf{44.67} $\mid$ $\times$ & \textcolor{VioletRed}{\textbf{0.112}} $\mid$ $\times$ \\\hline
\end{tabular}
\end{table*}

\subsection{Scene segmentation}

\begin{figure*}[htb]
    \centering
    \includegraphics[width=\textwidth]{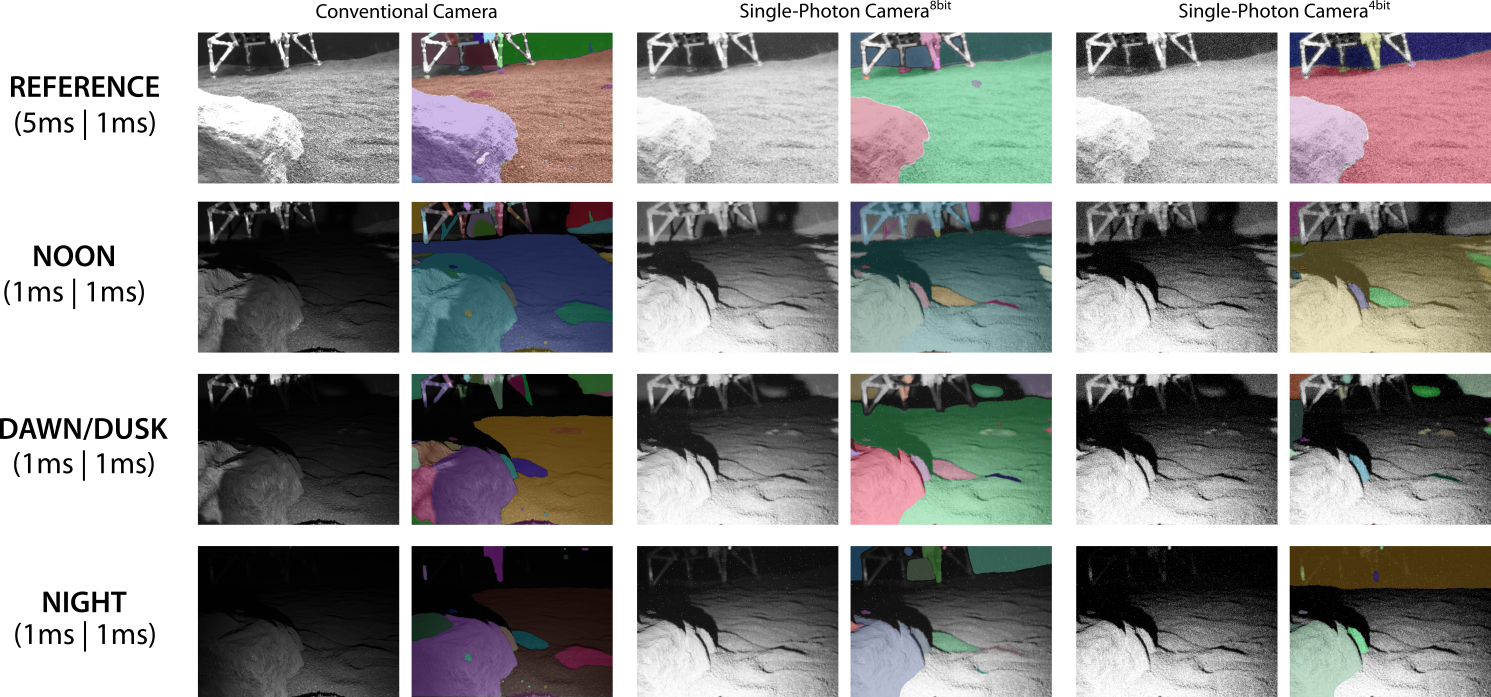}
    \caption{Segmented images resulted from running SAM \cite{kirillov2023segment} over conventional and single-photon images captured under different illumination scenarios. All masks are semantic-free and randomly colorized. Segmentations from 4-bit single-photon frames are also illustrated, to demonstrate the SPAD’s potential ability to reproduce comparable results even at reduced bit depths, despite models like SAM not being originally designed for low-dimensional data. Note that the equivalent exposure time at 4 bit-per-pixel would be 80$\mu$s.}
    \label{fig:segmentation}
\end{figure*}

Results from running the SAM ViT-H model over a single frame captured at different illumination conditions are shown in Figure~\ref{fig:segmentation}, with associated metrics reported in Table~\ref{tab:segmentation}. While conclusions from a single image are necessarily limited, this frame was selected for its challenging backlit conditions and the presence of both natural (foreground rock) and artificial (lander) objects, including what can be considered an artificial horizon line in the background.

{\setlength{\tabcolsep}{3pt}
\begin{table}[]
\small
\centering
\caption{Segmentation metrics obtained from the results presented in Figure~\ref{fig:segmentation}.}\vspace{-0.5em}
\begin{tabular}{ccccc}
\hline
Illumination & Camera config.  & Inference (s) & mIoU & \#Masks \\ \hline
\multirow{3}{*}{Reference}  & Conventional & 95.80 &  1.00 & \textbf{35} \\
                            & SPAD 8bit & 99.60  & 1.00 & 23 \\
                            & SPAD 4bit & \textbf{79.39} & 0.705  & 11 \\ \hline
                            
\multirow{3}{*}{Noon}       & Conventional & 89.71 &  \textbf{0.271} & \textbf{38} \\
                            & SPAD 8bit & \textbf{88.30}  & 0.265 & 21  \\
                            & SPAD 4bit & 119.12  & 0.191 & 15 \\ \hline
                            
\multirow{3}{*}{Dawn/Dusk}  & Conventional & 105.23 &  \textbf{0.231} & \textbf{29}  \\
                            & SPAD 8bit & \textbf{97.24}  & 0.226 & 27  \\
                            & SPAD 4bit & 100.22  & 0.131  & 21 \\ \hline

\multirow{3}{*}{Night}      & Conventional & \textbf{91.94} &  0.122 & \textbf{21}  \\
                            & SPAD 8bit & 98.27  & 0.265 & 19 \\
                            & SPAD 4bit & 92.48  & \textbf{0.3383}  &  9 \\ \hline
\end{tabular}
\label{tab:segmentation}
\end{table}
}

Under well-lit reference and noon scenarios, all camera configurations yield clear segmentations, with distinct object boundaries and color-coded masks. As illumination decreases, however, differences emerge. At 1\,ms exposure in low light (dawn/dusk and night), the conventional camera produces underexposed images with substantial detail loss, enabling reasonable foreground segmentation but failing in the background without longer exposures or post-processing. The SPAD camera maintains robust segmentation at 8-bit during dawn/dusk, but at night, it also struggles to capture most background elements. At 4-bit depth, performance becomes less consistent, sometimes preserving object shapes, but at other times fragmenting boundaries or missing objects entirely, even when they are visually discernible. This suggests that segmentation models like SAM could benefit from training on lower bit-depth data to fully exploit the resource-efficient capabilities of SPAD cameras in photon-scarce environments.

Most conditions beyond the reference scenario result in a significant decrease in mIoU for all configurations, with the conventional camera generally achieving slightly higher mIoU and detecting more masks at comparable inference times. Notably, at night, bit depth does not strictly correlate with mIoU: the 4-bit SPAD achieves a higher mIoU (0.338) than the conventional camera (0.122), indicating that performance rankings can invert in extremely low light. Inference time varies modestly across cameras and illumination conditions, with no clear link to segmentation accuracy.

While mIoU offers a convenient single-value measure, it has limitations under poor visual conditions. Low mIoU may stem from missed detections rather than poor pixel alignment, and differences in exposure or sensor characteristics can exaggerate performance gaps. For instance, at night, the conventional camera detects 21 masks with a very low mIoU, while the 4-bit SPAD detects only 9 masks but achieves a higher mIoU. This highlights the need to interpret mIoU alongside mask counts and qualitative inspection for a more comprehensive assessment of segmentation performance.

\subsection{Lander detection}

\begin{figure}[htb]
    \centering
    \includegraphics[width=0.9\linewidth]{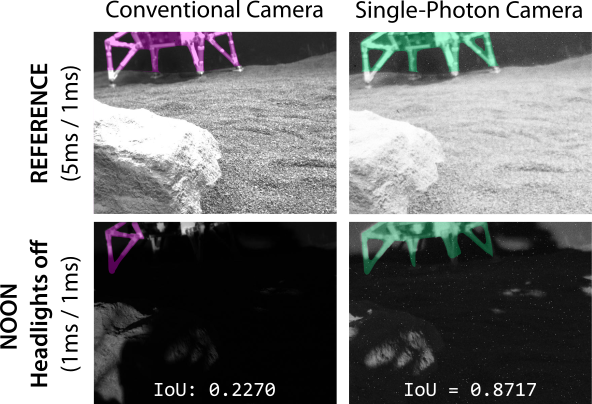}
    \caption{Results from extracting a SAM-based mask of the background lander structure on both conventional and single-photon camera images.}
    \label{fig:lander}
\end{figure}

Under well-lit conditions, both camera types have proven to capture the scene with high fidelity. To further test the extreme low-light performance of the single-photon camera, however, we wanted to evaluate segmentation results using the same image from the previous experiment, but acquired under noon conditions with the rover headlights switched off. The aim was to assess whether each camera could capture the full shape of the lander with sufficient quality for a SAM-based mask to detect it in its entirety. Figure~\ref{fig:lander} shows the results of this evaluation. Unlike the conventional camera, the SPAD camera successfully resolved most of the visible lander’s structure under these conditions.

\subsection{Visual Odometry}

\begin{figure}[ht]
    \centering
    \includegraphics[width=0.8\linewidth]{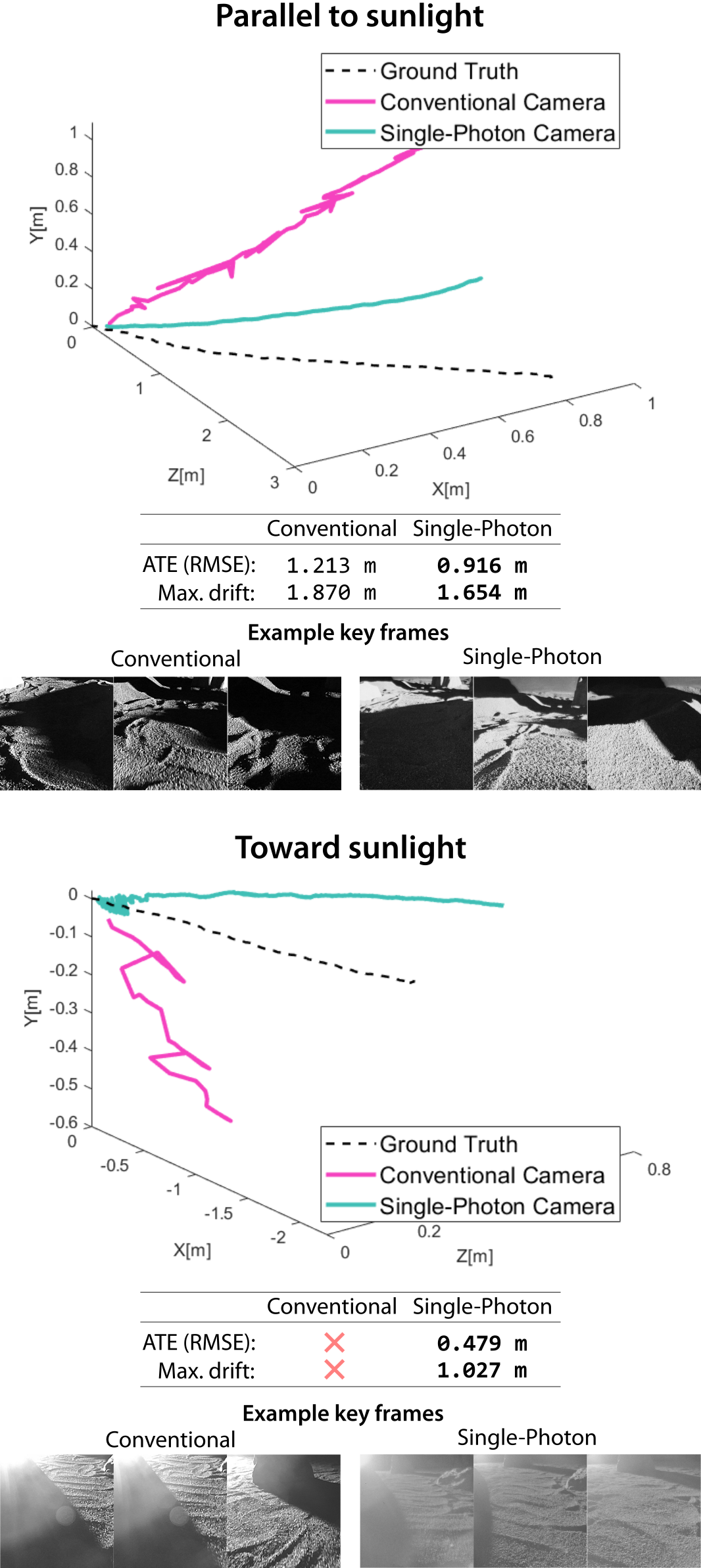}
    \caption{Results from running ORBSLAM3 over two types of trajectories: parallel to sunlight (top) and toward sunlight (bottom). Metrics are computed based on a first-pose alignment of both estimated and ground truth trajectories.}
    \label{fig:vo}
\end{figure}

Lastly, we evaluated the performance of both camera types in a localization task. Figure~\ref{fig:vo} compares the trajectories estimated by ORB-SLAM3 using images captured with both cameras under trajectories B (parallel to sunlight) and C (toward sunlight) of the SPICE-HL3 dataset. These images were captured during dawn/dusk conditions, with the rover’s headlights on and moving at a speed of 0.5 m/s. For trajectory D (away from sunlight), neither camera recorded sufficient image features for a successful localization, resulting in the most challenging scenario. Despite the extreme visual conditions---including complex lighting and the limited field of view of both cameras (conventional camera images were cropped to match the aspect ratio and field of view of the SPAD)---the SPAD camera consistently produced trajectories that more closely followed the ground truth. Note that when facing direct sunlight, conventional camera images were heavily affected by lens flares and oversaturation, which prevented enough features from being identified and tracked, making localization impossible. In contrast, the SPAD camera proved more reliable under strong illumination and shadow, demonstrating robust performance in HDR lighting conditions. These results underscore the advantage of single-photon imaging for visual odometry in environments with perceptually degraded conditions. As with the SAM model used in the scene segmentation evaluation, it is important to emphasize that pipelines such as ORB-SLAM3 are ill-suited for processing low-dimensional data, which is made available directly from the SPAD raw binary stream, making it impossible to fully assess the capabilities of the SPAD in its entirety.

\section{Conclusions}
\label{sec:conclusions}
Our work confirmed that SPAD cameras demonstrate clear advantages over conventional sensors in perceptually degraded environments, particularly those affected by extremely low illuminance or demanding sensitivity to an HDR scene. Under suboptimal illumination, conventional sensors require either longer exposures or rapid exposure adaptations, both of which can exacerbate noise in images captured from fast-moving platforms, such as those moving exceedingly fast, like spacecraft in orbit. This reduced SNR limits the effectiveness of downstream computational imaging methods. However, to match the performance and efficiency of established technologies under high photon flux conditions, further advancements are required. Recent developments, such as the integration of microlenses to improve pixel fill factor and photon detection efficiency, the addition of color filters, and the increasing sensor resolution, are promising steps toward addressing these limitations.

This evaluation is inherently constrained by the algorithms used, many of which assume 8-bit intensity data as input. To fully exploit the potential of SPADs in space and robotic applications, new algorithms must be designed to operate directly on the lower-dimensional high-throughput data SPADs produce, particularly their raw binary photon detection streams. Specialized preprocessing pipelines will also be essential for handling this data format efficiently. However, working directly with raw binary streams introduces bandwidth challenges; for many tasks, transmitting all recorded binary frames may not be neither necessary nor optimal. For example, excessively high frame rates can degrade performance in monocular visual odometry, where a minimum amount of inter-frame motion is required for a stable initialization. Adaptive frame-rate control could mitigate this issue while conserving data bandwidth. 

Adaptability is, indeed, one of the two main advantages of single-photon cameras, i.e., the ability to deliver tailored, size-efficient bit-depth images at any frame rate required by the task at hand. The other is their exceptional light sensitivity and speed. An underexplored feature of SPADs for imaging under these types of illumination conditions is their potential for ultra-fast HDR imaging, achieved by combining binary streams captured at different exposure settings in near real time.

Our results show that single-photon technology can deliver robust imaging in conditions where conventional sensors often fail, highlighting SPADs’ potential for autonomous robots operating in some of the most visually challenging planetary environments.







\section*{Acknowledgments}

This work was conducted under the project INSIGHT (PID2024-160373OB-C21) funded by the Spanish Ministry of Science, Innovation, and Universities, the Spanish State Research Agency, and the European Regional Development Fund (ERDF, UE). We extend our appreciation to Pi Imaging Technology for lending us the SPAD camera that made this study possible.


\footnotesize
\bibliography{references}

\end{document}